# Adaptive Problem-Solving for Large-Scale Scheduling Problems: A Case Study

**Jonathan Gratch**                                             GRATCH@ISI.EDU

*University of Southern California, Information Sciences Institute*
*4676 Admiralty Way, Marina del Rey, CA 90292, USA*

**Steve Chien**                                             STEVE.CHIEN@JPL.NASA.GOV

*Jet Propulsion Laboratory, California Institute of Technology*
*4800 Oak Grove Drive, M/S 525–3660, Pasadena, CA, 91109–8099*

## Abstract

Although most scheduling problems are NP-hard, domain specific techniques perform well in practice but are quite expensive to construct. In *adaptive problem-solving,* domain specific knowledge is acquired automatically for a general problem solver with a flexible control architecture. In this approach, a learning system explores a space of possible heuristic methods for one well-suited to the eccentricities of the given domain and problem distribution. In this article, we discuss an application of the approach to scheduling satellite communications. Using problem distributions based on actual mission requirements, our approach identifies strategies that not only decrease the amount of CPU time required to produce schedules, but also increase the percentage of problems that are solvable within computational resource limitations.

## 1.  Introduction

With the maturation of automated problem-solving research has come grudging abandonment of the search for "the" domain-independent problem solver. General problem-solving tasks like planning and scheduling are provably intractable. Although heuristic methods are effective in many practical situations, an ever growing body of work demonstrates the narrowness of specific heuristic strategies (e.g., Baker, 1994, Frost & Dechter, 1994, Kambhampati, Knoblock & Yang, 1995, Stone, Veloso & Blythe, 1994, Yang & Murray, 1994). Studies repeatedly show that a strategy that excels on one task can perform abysmally on others. These negative results do not entirely discredit domain-independent approaches, but suggest that considerable effort and expertise is required to find an acceptable combination of heuristic methods, a conjecture that is generally by published accounts of real-world implementations (e.g., Wilkins, 1988). The specificity of heuristic methods is especially troubling when we consider that problem-solving tasks frequently change over time. Thus, a heuristic problem solver may require expensive "tune-ups" as the character of the application changes.

*Adaptive problem solving* is a general method for reducing the cost of developing and maintaining effective heuristic problem solvers. Rather than forcing a developer to choose a specific heuristic strategy, an adaptive problem solver adjusts itself to the idiosyncrasies of an application. This can be seen as a natural extension of the principle of least commitment (Sacerdoti, 1977). When solving a problem, one should not commit to a particular solution path until one has information to distinguish that path from the alternatives. Likewise, when faced with an entire distribution of problems, it makes sense to avoid committing to a particular heuristic strategy until one can make an informed decision on which strategy performs better on the distribution. An adaptive problem solver embodies a space of heuristic methods, and only settles on a particular combination of these methods after a period of adaptation, during which the system automatically acquires information about the particular distribution of problems associated with the intended application.

In previous articles, Gratch and DeJong have presented a formal characterization of adaptive problem solving and developed a general method for transforming a standard problem solver into an adaptive one (Gratch & DeJong, 1992, Gratch & DeJong, 1996). The primary purpose of this article is twofold: to illustrate the efficacy of learning approaches for solving real-world problem solving tasks, and to build empirical support for the the specific learning approach we advocate. After reviewing the basic method, we describe its application to the development of a large-scale scheduling system for the National Aeronautics and Space Administration (NASA). We applied the adaptive problem solving approach to a scheduling system developed by a separate research group, and without knowledge of our adaptive techniques. The scheduler included an expert-crafted scheduling strategy to achieve efficient scheduling performance. By automatically adapting this scheduling system to the distribution of scheduling problems, the adaptive approach resulted in a significant improvement in scheduling performance over an expert strategy: the best adaptation found by machine learning exhibited a seventy percent improvement in scheduling performance (the average learned strategy resulted in a fifty percent improvement).

## 2. Adaptive Problem Solving

An adaptive problem solver defers the selection of a heuristic strategy until some information can be gathered about their performance over the specific distribution of tasks. The need for such an approach is predicated on the claim that it is difficult to identify an effective heuristic strategy *a priori*. While this claim is by no means proven, there is considerable evidence that, at least for the class of heuristics that have been proposed till now, no one collection of heuristic methods will suffice. For example, Kambhampati, Knoblock, and Yang (1995) illustrate how planning heuristics embody *design tradeoffs* — heuristics that reduce the size of search space typically increase the cost at each node, and vice versa — and that the desired tradeoff varies with different domains. Similar observations have been made in the context of constraint satisfaction problems (Baker, 1994, Frost & Dechter, 1994). This inherent difficulty in recognizing the worth (or lack of worth) of control knowledge has been termed the *utility problem* (Minton, 1988) and has been studied extensively in the machine learning community (Gratch & DeJong, 1992, Greiner & Jurisca, 1992, Holder, 1992, Subramanian & Hunter, 1992). In our case the utility problem is determining the worth of a heuristic strategy for specific problem distribution.

### 2.1  Formulation of Adaptive problem solving

Before discussing approaches to adaptive problem solving, we formally state the common definition of the task (as proposed by Gratch & DeJong, 1992, Greiner & Jurisca, 1992, Laird, 1992, Subramanian & Hunter, 1992). Adaptive problem solving requires a flexible problem solver, meaning the problem solver possesses control decisions that may be resolved in alternative ways. Given a flexible problem solver, $PS$, with several control points, $CP_1...CP_n$ (where each control point $CP_i$ corresponds to a particular control decision), and a set of alternative heuristic methods for each control point, $\{M_{i,1}...M_{i,k},\}$,[1] a *control strategy* defines a specific method for every control point (e.g., $STRAT = <M_{1,3}, M_{2,6}, M_{3,1},...>$). A control strategy determines the overall behavior of the problem solver. Let $PS_{STRAT}$ be the problem solver operating under a particular control strategy.

The quality of a problem solving strategy is defined in terms of the decision-theoretic notion of expected utility. Let $U(PS_{STRAT}, d)$, be a real valued *utility function* that is a measure of the goodness of the behavior of the problem solver on a specific problem $d$. More generally, expected utility can be defined formally over a distribution of problems $D$:

$$E_D[U(PS_{STRAT})] = \sum_{d \in D} U(PS_{STRAT}, d) \times probability(d)$$

The goal of adaptive problem solving can be expressed as: given a problem distribution $D$, find some control strategy in the space of possible strategies that maximizes the expected utility of the problem solver. For example, in the PRODIGY planning system (Minton, 1988), control points include: how to select an operator to use to achieve the goal; how to select variable bindings to instantiate the operator; etc. A method for the operator choice control point might be a set of control rules to determine which operators to use to achieve various goals. A strategy for PRODIGY would be a set of control rules and default methods for every control point (e.g., one for operator choice, one for binding choice, etc.). Utility might be defined as a function of the time to construct a plan for a given planning problem.

## 2.2 Approaches to Adaptive Problem Solving

Three potentially complementary approaches to adaptive problem solving have been discussed in the literature. The first, what we call a *syntactic approach*, is to preprocess a problem-solving domain into a more efficient form, based solely on the domain's syntactic structure. For example, Etzioni's STATIC system analyzes a portion of a planing domain's deductive closure to conjecture a set of search control heuristics (Etzioni, 1990). Dechter and Pearl describe a class of constraint satisfaction techniques that preprocess a general class of problems into a more efficient form (Dechter & Pearl, 1987). More recent work has focused on recognizing those structural properties that influence the effectiveness of different heuristic methods (Frost & Dechter, 1994, Kambhampati, Knoblock & Yang. 1995, Stone, Veloso & Blythe, 1994). The goal of this research is to provide a problem solver with what is essentially a big lookup table, specifying which heuristic strategy to use based on some easily recognizable syntactic features of a domain. While this later approach seems promising, work in this area is still preliminary and has focused primarily on artificial applications. The disadvantage of purely syntactic techniques is that they ignore a potentially important source of information, the distribution of problems. Furthermore, current syntactic approaches to this problem are specific to a particular, often unarticulated, utility function (usually problem-solving cost). For example, allowing the utility function to be a user specified parameter would require a significant and problematic extension of these methods.

The second approach, which we call a *generative approach,* is to generate custom-made heuristics in response to careful, automatic, analysis of past problem-solving attempts. Generative ap-proaches consider not only the structure of the domain, but also structures that arise from the problem solver interacting with specific problems from the domain. This approach is exemplified by SOAR (Laird, Rosenbloom & Newell, 1986) and PRODIGY/EBL (Minton, 1988). These techniques analyze past problem-solving traces and conjectures heuristic control rules in response to specific problem-solving inefficiencies. Such approaches can effectively exploit the idiosyncratic structure of a domain through this careful analysis. The limitation of such approaches is that they have typically focused on generating heuristics in response to particular problems and have not well addressed the issue of adapting to a distribution of problems[2]. Furthermore, as with the syntactic approaches, thus far they have been directed towards a specific utility function.

---

1. Note that a method may consist of smaller elements so that a method may be a set of control rules or a combination of heuristics.

The final approach we call the *statistical approach.* These techniques explicitly reason about performance of different heuristic strategies across the distribution of problems. These are generally statistical generate-and-test approaches that estimated the average performance of different heuristics from a random set of training examples, and explore an explicit space of heuristics with greedy search techniques. Examples of such systems are COMPOSER (Gratch & DeJong, 1992), PALO (Greiner & Jurisca, 1992), and the statistical component of MULTI-TAC (Minton, 1993). Similar approaches have also been investigated in the operations research community (Yakowitz & Lugosi, 1990). These techniques are easy to use, apply to a variety of domains and utility functions, and can provide strong statistical guarantees about their performance. They are limited, however, as they are computationally expensive, require many training examples to identify a strategy, and face problems with local maxima. Furthermore, they typically leave it to the user to conjecture the space of heuristic methods (see Minton, 1993 for a notable exception).

In this article, we adopt the statistical approach to adaptive problem solving due to its generality and ease of use. In particular we use the COMPOSER technique for adaptive problem solving (Gratch & DeJong, 1992, Gratch & DeJong, 1996), which is reviewed in the next section. Our implementation incorporates some novel features to address the computational expense of the method. Ideally, however, an adaptive problem solver would incorporate some form of each of these methods. To this end we are investigating how to incorporate other methods of adaptation in our current research.

## 3. COMPOSER

COMPOSER embodies a statistical approach to adaptive problem solving. To turn a problem solver into an adaptive problem solver, the developer is required to specify a utility function, a representative sample of training problems, and a space of possible heuristic strategies. COMPOSER then adapts the problem solver by exploring the space of heuristics via statistical hillclimbing search. The search space is defined in terms of a *transformation generator* which takes a strategy and generates a set of transformations to it. For example, one simple transformation generator just returns all single method modifications to a given strategy. Thus a transformation generator defines both a space of possible heuristic strategies and the non-deterministic order in which this space may be searched. COMPOSER's overall approach is one of generate and test hillclimbing. Given an initial problem solver, the transformation generator returns a set of possible transformations to its control strategy. These are statistically evaluated over the expected distribution of problems. A transformation is adopted if it increases the expected performance of solving problems over that distribution. The generator then constructs a set of transformations to this new strategy and so on, climbing the gradient of expected utility values.

---

2. While generative approaches can be trained on a problem distribution, learning typically occurs only within the context of a single problem. These systems will often learn knowledge which is helpful in a particular problem but decreases utility overall, necessitating the use of utility analysis techniques.

Formally, COMPOSER takes an initial problem solver, $PS_0$, and identifies a sequence of problem solvers, $PS_0$, $PS_1$, ... where each subsequent $PS$ has higher expected utility with probability $1-\delta$ (where $\delta > 0$ is some user–specified constant). The transformation generator, $TG$, is a function that takes a problem solver and returns a set of candidate changes. $Apply(t, PS)$ is a function that takes a transformation, $t \in TG(PS)$ and a problem solver and returns a new problem solver that is the result of transforming $PS$ with $t$. Let $U_j(PS)$ denote the utility of $PS$ on problem $j$. The change in utility that a transformation provides for the $j$th problem, called the *incremental utility* of a transformation, is denoted by $\Delta U_j(t/PS)$. This is the difference in utility between solving the problem with and without the transformation. COMPOSER finds a problem solver with high expected utility by identifying transformations with positive expected incremental utility. The expected incremental utility is estimated by averaging a sample of randomly drawn incremental utility values. Given a sample of $n$ values, the average of that sample is denoted by $\overline{\Delta U}_n(t/PS)$. The likely difference between the average and the true expected incremental utility depends on the variance of the distribution, estimated from a sample by the *sample variance $S_n^2(t|PS)$*, and the size of the sample, $n$. COMPOSER provides a statistical technique for determining when sufficient examples have been gathered to decide, with error $\delta$, that the expected incremental utility of a transformation is positive or negative. Because COMPOSER presumes that the relevant distributions are normally distributed, COMPOSER requires at that each estimate of incremental utility be based on a minimum number of samples $n_0$ to be determined for each application. The algorithm is summarized in Figure 1.

COMPOSER's technique is applicable in cases where the following conditions apply:

1. The control strategy space can be structured to facilitate hillclimbing search. In general, the space of such strategies is so large as to make exhaustive search intractable. COMPOSER requires a transformation generator that structures this space into a sequence of search steps, with relatively few transformations at each step. In Section 5.1 we discuss some techniques for incorporating domain specific information into the structuring of the control strategy space.

2. There is a large supply of representative training problems so that an adequate sampling of problems can be used to estimate expected utility for various control strategies.

3. Problems can be solved with a sufficiently low cost in resources so that estimating expected utility is feasible.

4. There is sufficient regularity in the domain such that the cost of learning a good strategy can be amortized over the gains in solving many problems.

## 4. The Deep Space Network

The Deep Space Network (DSN) is a multi-national collection of ground-based radio antennas responsible for maintaining communications with research satellites and deep space probes. DSN Operations is responsible for scheduling communications for a large and growing number of spacecraft. This already complex scheduling problem is becoming more challenging each year as budgetary pressures limit the construction of new antennas. As a result, DSN Operations has turned

---

**Given:** $PS_{old}$, $TG(\cdot)$, $\delta$, examples, $n_0$

[1] $PS := PS_{old}$; $T := TG(PS)$; $n := 0$; $i := 0$; $\alpha := Bound(\delta, |T|)$;

[3]      Repeat     {Find next transformation}

[2] While $T \neq \varnothing$ and $i < |\text{examples}|$ do    {Hillclimb as long as there is data and possible transformations}

[4]         $n := n+1$; $i := i+1$; step–taken := FALSE;

[5]         $\forall \tau \in$ T: Get $\Delta U_i(\tau/PS)$    {Observe incremental utility values for $i$th problem}

[6]         $significant := \left\{ \tau \in T : n \geq n_0 \text{ and } \dfrac{S_n^2(\tau|PS)}{\left[\overline{\Delta U}_n(\tau|PS)\right]^2} < \dfrac{n}{\left[Q(\alpha)\right]^2} \right\}$

         {Collect all transformations that have reached statistical significance.}

[7]         $T := T - \left[ \tau \in significant : \overline{\Delta U}_n(\tau|PS) < 0 \right]$    {Discard trans. that decrease expeced utility}

[8]         If $\exists \tau \in significant : \overline{\Delta U}_n(\tau|PS) > 0$ Then   {Adopt $\tau$ that most increases expected utility}

[9]            $PS = Apply(x \in significant : \forall y \in significant \left[\overline{\Delta U}_n(x|PS) > \overline{\Delta U}_n(y|PS)\right], PS)$

[10]         $T := TG(PS)$; $n := 0$; $\alpha := Bound(\delta, |T|)$; step–taken :=TRUE;

[11]     Until step–taken or $T = \varnothing$ or $i = |\text{examples}|$;

**Return:** $PS$

$$Bound(\delta, |T|) := \frac{\delta}{|T|}, \qquad Q(\alpha) := x \ where \int_x^\infty \left(1/\sqrt{2\pi}\right)e^{-0.5y^2}dy = \frac{\alpha}{2}$$

Figure 1: The COMPOSER algorithm

---

increasingly towards intelligent scheduling techniques as a way of increasing the efficiency of network utilization. As part of this ongoing effort, the Jet Propulsion Laboratory (JPL) has been given the responsibility of automating the scheduling of the 26-meter sub-net; a collection of 26-meter antennas at Goldstone, CA, Canberra, Australia and Madrid, Spain.

In this section we discuss the application of adaptive problem-solving techniques to the development of a prototype system for automated scheduling of the 26-meter sub-net. We first discuss the development of the basic scheduling system and then discuss how adaptive problem solving enhanced the scheduler's effectiveness.

### 4.1 The Scheduling Problem

Scheduling the DSN 26-meter subnet can be viewed as a large constraint satisfaction problem. Each satellite has a set of constraints, called project requirements, that define its communication needs. A typical project specifies three generic requirements: the minimum and maximum number of communication events required in a fixed period of time; the minimum and maximum duration for these communication events; and the minimum and maximum allowable gap between communication events. For example, Nimbus-7, a meteorological satellite, must have at least four 15-minute communication slots per day, and these slots cannot be greater than five hours apart. Project requirements are determined by the project managers and tend to be invariant across the lifetime of the spacecraft.

In addition to project requirements, there are constraints associated with the various antennas. First, antennas are a limited resource – two satellites cannot communicate with a given antenna at the same time. Second, a satellite can only communicate with a given antenna at certain times, depending on when its orbit brings it within view of the antenna. Finally, antennas undergo routine maintenance and cannot communicate with any satellite during these times.

Scheduling is done on a weekly basis. A weekly scheduling problem is defined by three elements: (1) the set of satellites to be scheduled, (2) the constraints associated with each satellite, and (3) a set of *time periods* specifying all temporal intervals when a satellite can legally communicate with an antenna for that week. Each time period is a tuple specifying a satellite, a communication time interval, and an antenna, where (1) the time interval must satisfy the communication duration constraints for the satellite, (2) the satellite must be in view of the antenna during this interval. Antenna maintenance is treated as a project with time periods and constraints. Two time periods conflict if they use the same antenna and overlap in temporal extent. A valid schedule specifies a non-conflicting subset of all possible time periods where each project's requirements are satisfied.

The automated scheduler must generate schedules quickly as scheduling problems are frequently over-constrained (i.e., the project constraints combined with the allowable time periods produces a set of constraints which is unsatisfiable). When this occurs, DSN Operations must go through a complex cycle of negotiating with project managers to reduce their requirements. A goal of automated scheduling is to provide a system with relatively quick response time so that a human user may interact with the scheduler and perform "what if" reasoning to assist in this negotiation process. Ultimately, the goal is to automate this negotiation process as well, which will place even greater demands on scheduler response time (Chien & Gratch, 1994). For these reasons, the focus of development is upon heuristic techniques that do not necessarily uncover the optimal schedule, but rather produce adequate schedules quickly.

### 4.2 The LR-26 Scheduler

LR-26 is a heuristic scheduling approach to DSN scheduling being developed at the Jet Propulsion Laboratory (Bell & Gratch, 1993).[3] LR-26 is based on a 0–1 integer linear programming formulation of the scheduling problem (Taha, 1982). Scheduling is cast as the problem of finding an assignment to integer variables that maximizes the value of some objective function subject to a set of linear constraints. In particular, time periods are treated as 0-1 integer variables: 0 (or OUT) if the time period is excluded from the schedule; 1 (or IN) if it is included. The objective is to maximize the number of time periods in the schedule and the solution must satisfy the project requirements and antenna constraints (expressed as sets of linear inequalities). A typical scheduling problem under this formulation has 700 variables and 1300 constraints.

In operations research, integer programs are solved by a variety of techniques including branch-and-bound search, the gomory method (Kwak & Schniederjans, 1987), and Lagrangian relaxation (Fisher, 1981). In artificial intelligence such problems are generally solved by constraint propagation search techniques (e.g., Dechter, 1992, Mackworth, 1992). To address the complexity of the scheduling problem LR-26 uses a hybrid approach that combines Lagrangian relaxation with constraint propagation search. Lagrangian relaxation is a divide-and-conquer method which, given a decomposition of the scheduling problem into a set of easier sub-problems, coerces the sub-problems to be solved in such a way that they frequently result in a global solution. One specifies a problem decomposition by identifying a subset of problem constraints that, if removed, result in one or more independent and computationally easy sub-problems.[4] These problematic constraints are "relaxed," meaning they no longer act as constraints but instead are added to the objective function in such a way that (1) there is incentive to satisfying these relaxed constraints when solving the sub-problems and, (2) the best solution to the relaxed problem, if it satisfies all relaxed constraints, is guaranteed to be the best solution to the original problem. Furthermore, this relaxed objective function is parameterized by a set of weights (one for each relaxed constraint). By systematically changing these weights (thereby modulating the incentives for satisfying relaxed constraints) a global solution can often be found. Even if this weight search does not produce a global solution, it can make the solution to the sub-problems sufficiently close to a global solution that a global solution can be discovered with substantially reduced constraint propagation search.

---

3. LR-26 stands for the Lagrangian Relaxation approach to scheduling the 26-meter sub-net.

In the DSN domain, the scheduling problem is decomposed by scheduling each antenna independently. Specifically, the constraints associated with the complete problem can be divided into two groups: those that refer to a single antenna, and those that mention multiple antennas. The later are relaxed and the resulting single-antenna sub-problems can be solved in time linear in the number of time periods associated with that antenna (see below). LR-26 solves the complete problem by first trying to coerce a global solution by performing a search in the space of weights and then, if that fails to produce a solution, resorting to constraint propagation search in the space of possible schedules.

### 4.2.1 SCHEDULES

We now describe the formalization of the problem. Let $P$ be a set of projects, $A$ a set of antennas, $M = \{0,...,10080\}$, and $V$ be an enumeration, $V=\{0, 1, *\}$, denoting whether a time period is excluded from the schedule (0), included (1), or uncommitted. Note that $P$, $A$, and $M$, are specified in advance and $V$ is to be determined by the scheduler and is initially always uncommitted. Let $S \subseteq P \times A \times M \times M \times V$ denote the set of possible time periods for a week, where a given time period specifies a project, antenna and the start and end of the communication event, respectively. For a given $s \in S$, we define $project(s)$, $antenna(s)$, $start(s)$, $end(s)$, and $value(s)$ to denote the corresponding elements of $s$. We also define $length(s) = end(s) - start(s)$ to simplify some subsequent notation.

A *ground schedule* is an assignment of 0 (excluded) or 1 (included) to each time period in $S$. This can be seen as the application to $S$ of some function $G$ that maps each element of $S$ to 0 or 1. We denote this by $S^G$. A *partial schedule* refers to a schedule with only a subset of its time periods committed, which we denote via some mapping function $M$ that maps elements of $S$ to 0, 1, or *. A partial schedule corresponds to a set of possible ground schedules (i.e., those that result from forcing each uncommitted time period either in or out of the schedule). We denote this by $S^M$. We define a particular partial schedule $S^0$ to denote the completely uncommitted partial schedule (with all time periods assigned a value of *).

---

4. A problem consists of *independent* sub-problems it the global objective function can be maximized by finding some maximal solution for each sub-problem in isolation.

## 4.2.2 CONSTRAINTS

The scheduler must identify some ground schedule that satisfies a set of project and antenna constraints, which we now formalize.

**Project Requirements.** Each project $p_n \in P$ has associated with it a set of constraints called project requirements. All constraints are processed and translated into simple linear inequalities over elements of $S$. The complete set of project requirements, denoted $PR$, is the union of the requirements from each individual projects. Each requirement can be expressed as integer linear inequality:

$$pr_j \in PR \equiv \sum_{s_i \in S} a_{i,j} \cdot value(s_i) \geq b_j \ \text{ or } \ \sum_{s_i \in S} a_{i,j} \cdot value(s_i) \geq b_j$$

where $a_i$ represents a weighting factor indicating the degree to which the $i$th time period (if included) contributes to satisfying a particular requirement. For example, the requirement that a project, $p$, must have at least 100 minutes of communication time in a week is expressed:

$$\sum_{s \in S} [I(project(s) = p) \cdot length(s)] \cdot value(s) \geq 100.$$

Where $I(project(s))$ equals one if $s$ belongs to that project; otherwise zero. Note that time periods with zero weight play no role and are not explicitly mentioned in the actual constraint representation.

Constraints on the length of individual time periods are represented similarly:

$$length(s) \geq 15$$

For efficiency, however, time periods which do not satisfy these unary inequalities are simply eliminated from $S$ in a preprocessing step.[5]

**Antenna Constraints.** Each of the three antennas has the constraint that no two projects can use the antenna at the same time. This can be translated into a set of linear inequalities $AC_a$, for each antenna $a$ as follows:

$$AC_a = \{s_i + s_j \leq 1 \mid s_i \neq s_j \wedge antenna(s_i) = antenna(s_j) = a \ \wedge$$
$$[start(s_i)..end(s_i)] \cap [start(s_j)..end(s_j)] \neq \varnothing\}$$

## 4.2.3 PROBLEM FORMULATION

The scheduling objective used by LR-26 is to find some ground schedule, denoted by $S^*$, that maximizes the number of time periods in the schedule subject to the project and antenna constraints:[6]

**Problem:** *DSN*

Find: $\quad S^* = \arg\max\limits_{S^G \in s^0} \left\{ Z^G = \sum\limits_{s \in S^s} value(s) \right\}$ (1)

Subject to: $\quad AC_1 \cup AC_2 \cup AC_3 \cup PR$

---

5. Note that this is an inherent limitation in the formalization as the scheduler cannot entertain variable length communication events – communication events must be discretized into a finite set of fixed length intervals.

where $Z^G$ is the value of the objective function for some ground schedule and "arg max" denotes the argument that leads to the maximum.

With Lagrangian relaxation, certain constraints are folded into the objective function in a standardized fashion. The intuition is to add some factor into the objective function that is negative iff the relaxed constraint is unsatisfied. If a constraint is of the form $\Sigma a_i s_i \geq b$, then $u[\Sigma a_i s_i - b]$ is added to the objective function, where $u$ is a non-negative weighting factor. Likewise, if the constraint is of the form $\Sigma a_i s_i \leq b$, then $u[b - \Sigma a_i s_i]$ is added. In L$_{R}$-26, only project requirements are relaxed:

---

**Problem:**      $DSN(u)$

    Find:                                                          (2)

$$S^*(u) =$$
$$\arg \max_{S^G \in S^0} \left\{ Z^G(u) = Z^G + \sum_{PR_{\leq}} u_j \left[ \sum_{s_i \in S^G} a_{ij} \cdot value(s_i) - b_j \right] + \sum_{PR_{\geq}} u_j \left[ b_j - \sum_{s_i \in S^G} a_{ij} \cdot value(s_i) \right] \right\}$$

    Subject to:       $AC_1 \cup AC_2 \cup AC_3$

---

where $Z_s(u)$ is the relaxed objective function and $u$ is a vector of non-negative weights of length $|PR|$ (one for each relaxed constraint). Note that this defines a space of relaxed solutions that depend on the weight vector $u$. Let $Z^*$ denote the value of the optimal solution of the original problem (Definition 1), and let $Z^*(u)$ denote value of the optimal solution to the relaxed problem (Definition 2) for a particular weight vector $u$. For any weight vector $u$, $Z^*(u)$ can be shown to be an upper bound on the value of $Z^*$. Thus, if a relaxed solution satisfies all of the original problem constraints, it is guaranteed to be the optimal solution to the original problem. Lagrangian relaxation proceeds by incrementally tightening this upper bound (by adjusting the weight vector) in the hope of identifying a global solution. A global solution cannot always be identified in this manner, so a complete scheduler must combine Lagrangian relaxation with some form of search.

### 4.2.4   SEARCH

If a solution cannot be found through weight adjustment, L$_{R}$–26 resorts to basic refinement search (Kambhampati, Knoblock & Yang, 1995) (or split-and-prune search (Dechter & Pearl, 1987)) in the space of partial schedules. In this search paradigm a partial schedule is recursively refined (split) into a set of more specific partial schedules. In the context of the DSN scheduling problem, refinement corresponds to forcing uncommitted time periods in or out of the schedule. A partial schedule would be pruned if all of its ground schedules violate the constraints. The scheduler is applied recursively to each refined partial schedule until some satisfactory ground schedule is found or all schedules are pruned.

Each refinement is further refined by propagating the local consequence of new commitment. After a variable is set to a particular value, each individual constraint which references that variable is analyzed to determine which time period would be forced in or out of the schedule as a result of the assignment. L$_{R}$–26 performs only partial constraint propagation, because complete propagation is computationally expensive. Specifically, if constraint C1 references time periods $s_2$, $s_4$ and $s_5$, and $s_2$ is assigned a value, LR–26 analyzes C1 to see if the new assignment determines the value of $s_4$ and/or $s_5$. If, for example, $s_4$ is constrained to take on a particular value, this triggers analysis of all constraints which contain $s_4$. This can be viewed as performing arc–consistency (Dechter, 1992). During the constraint propagation it may be possible to show that the refinement contains no valid ground schedule. In this case the partial schedule may be pruned from the search.

---

6. This might correspond to a desire to maintain maximum downlink flexibility.

LR-26 augments this basic refinement search with Lagrangian relaxation to heuristically reduce the combinatorics of the problem. The difficulty with refinement search is that it may have to perform considerable (and poorly directed) search through a tree of refinements to identify a *single* satisficing solution. If an optimal solution is sought, every leaf of this search tree must be examined.[7] In contrast, by searching through the space of relaxed solutions to a partial schedule, one can sometimes identify the *best* schedule without any refinement search. Even when this is not possible, Lagrangian relaxation heuristically identifies a small set of problematic constraints, focusing the subsequent refinement search. Thus, by performing some search in the space of relaxed solutions at each step, the augmented search method can significantly reduce both the depth and breadth of refinement search.

The augmented procedure works to the extent that it can efficiently solve relaxed solutions, ideally allowing the algorithm to explore several points in the space of weight vectors in each step of the refinement search. LR-26 solves relaxed problems in linear time, $O(|AC_1 \cup AC_2 \cup AC_3|)$. To see this, note that each time period appears on exactly one antenna. Thus, $Z_s(u)$ can be broken into the sum of three objective functions, each containing only the time periods associated with a particular antenna. Furthermore, the relaxed objective function can be re–expressed as the weighted sum of each of the time periods on that antenna, and the unrelaxed constraints are simple pair–wise exclusion constraints between individual time periods. Combine this with the fact that time periods are partially ordered by their start time and the problem simplifies to identifying some non–exclusive sequence of time periods with the maximum cumulative weight. This is easily formulated and solved as a dynamic programming problem (see Bell & Gratch, 1993 for more details).

The augmented refinement search performed by LR-26 is summarized in Figure 2

### 4.2.5 Performance Tradeoffs

Perhaps the most difficult decisions in constructing the scheduler involve how to flesh out the details of steps 1,2, 3, and 4. The constraint satisfaction and operations research literatures have proposed many heuristic methods for these steps. Unfortunately, due to their heuristic nature, it is not clear what combination of methods best suits this scheduling problem. The power of a heuristic method depends on subtle factors that are difficult to assess in advance. Additionally, when considering multiple methods, one has to consider interactions between methods.

In LR-26 a key interaction arises in the tradeoff between the amount of weight vector search vs. refinement search performed by the scheduler (as determined by Step 2). At each step in the refinement search, the scheduler has the opportunity to search in the space of relaxed solutions. Spending more effort in this weight search can reduce the amount of subsequent refinement search. But at some point the savings in reduced refinement search may be overwhelmed by the cost of performing the

---

7. Partial schedules may also be pruned, as in branch-and-bound search, if they can be shown to contain lower value solutions that other partial schedules. In practice LR-26 is run in a *satisficing mode*, meaning that search terminates as soon as a ground schedule is found (not necessarily optimal) that satisfies all of the problem constraints.

**LR-26 Scheduler**

> *Agenda* := *{S⁰}*;
> While *Agenda* ≠ ∅

(1) Select some partial schedule $S \in Agenda$; $Agenda := Agenda - \{S\}$
(2) Weight search for some $S*(u) \in S$;
IF $S*(u)$ satisfies the project requirements ($PR$) Then
> *Return S\*(u)*;
Else
(3) Select constraint $c \in PR$ not satisfied by $S*(u)$;
(4) Refine $S$ into $\{S^i\}$, such that each $S^G \in S^i$ satisfies $c$
> and $\cup\{S^i\} = S$;
Perform constraint propagation on each $S^i$
> *Agenda* := *Agenda*$\cup\{S^i\}$;

Figure 2: The basic LR-26 refinement search method.

weight search. This is a classic example of the utility problem, and it is difficult to see how best to resolve the tradeoff without intimate knowledge of the form and distribution of scheduling problems.

Another important issue for improving scheduling efficiency is the choice of heuristic methods for controlling the direction of refinement search (as determined by steps 1, 3, and 4). Often these methods are stated as general principles (e.g., "first instantiate variables that maximally constrain the rest of the search space", Dechter, 1992, p. 277) and there may be many ways to realize them in a particular scheduler and domain. Furthermore, there are almost certainly interactions between methods used at different control points that make it difficult to construct a good overall strategy.

These tradeoffs conspire to make manual development and evaluation of heuristics a tedious, uncertain, and time consuming task that requires significant knowledge about the domain and scheduler. In the case of LR-26, its initial control strategy was identified by hand, requiring a significant cycle of trial-and-error evaluation by the developer over a small number of artificial problems. Even with this effort, the resulting scheduler is still expensive to use, motivating us to try adaptive techniques.

## 5. Adaptive Problem Solving for The Deep Space Network

We developed an adaptive version of the scheduler, *Adaptive LR-26,* in an attempt to improve its performance.[8] Rather than committing on a particular combination of heuristic strategies, Adaptive LR-26 embodies an adaptive problem solving solution. The scheduler is provided a variety of heuristic methods, and, after a period of adaptation, settles on a particular combination of heuristics that suits the actual distribution of scheduling problems for this domain.

To perform adaptive problem solving, we must formally specify three things: a transformation generator that defines the space of legal heuristic control strategies; a utility function that captures our preferences over strategies in the control grammar; and a representative sample of training problems. We describe each of these elements as they relate to the DSN scheduling problem.

### 5.1 Transformation Generator

The description of LR-26 in Figure 2 highlights four points of non-determinism with respect to how the scheduler performs its refinement search. To fully instantiate the scheduler we must specify: a way of ordering elements on the agenda, a weight search method, a method for selecting a constraint, and a method for generating a spanning set of refinements that satisfy the constraint. The alternative ways for resolving these four decisions are specified by a *control grammar*, which we now describe. The grammar defines the space of legal search control strategies available to the adaptive problem solver.

---

8. This system has also been referred to by the name DSN-COMPOSER (Gratch, Chien & DeJong, 1993).

### 5.1.1 SELECT SOME PARTIAL SCHEDULE

The first decision in the refinement search is to choose some partial schedule from the agenda. This selection policy defines the character of the search. Maintaining the agenda as a stack implements depth-first search. Sorting the agenda by some value function implements a best-first search. In Adaptive LR-26 we restrict the space of methods to variants of depth-first search. Each time a set of refinements is created (Decision 4), they are added to the front of the agenda. Search always proceeds by expanding the first partial schedule on the agenda. Heuristics act by ordering refinements before they are added to the agenda. The grammar specifies several ordering heuristics, sometimes called *value ordering heuristics,* or *look–ahead schemes* in the constraint propagation literature (Dechter, 1992, Mackworth, 1992). As these methods are entertained during refinement construction, their detailed description is delayed until that section.

Look-ahead schemes decide how to refine partial schedules. Look-back schemes handle the reverse decision of what to do whenever the scheduler encounters a dead end and must backtrack to another partial schedule. Standard depth-first search performs *chronological backtracking*, backing up to the most recent decision. The constraint satisfaction literature has explored several heuristic alternatives to this simple strategy, including backjumping (Gaschnig, 1979), backmarking (Haralick & Elliott, 1980), dynamic backtracking (Ginsberg, 1993), and dependency-directed backtracking (Stallman & Sussman, 1977) (see Backer & Baker, 1994, and Frost and Dechter, 1994, for a recent evaluation of these methods). We are currently investigating look-back schemes for the control grammar but they will not be discussed in this article.

### 5.1.2 SEARCH FOR SOME RELAXED SOLUTION

The next dimension of flexibility is in weight-adjusting methods to search the space of possible relaxed solutions for a given partial schedule. The general goal of the weight search is to find a relaxed solution that is closest to the true solution in the sense that as many constraints are satisfied as possible. This can be achieved by minimizing the value of $Z^*(u)$ with respect to $u$. The most popular method of searching this space is called *subgradient-optimization* (Fisher, 1981). This is a standard optimization method that repeatedly changes the current $u$ in the direction that most decreases $Z^*(u)$. Thus at step $i$, $u_{i+1} = u_i + t_i d_i$ where $t_i$ is a step size and $d_i$ is a directional vector in the weight space. The method is expensive but it is guaranteed to converge to the minimum $Z^*(u)$ under certain conditions (Held & Karp, 1970). A less expensive technique, but without the convergence guarantee, is to consider only one weight at a time when finding an improving direction. Thus $u_{i+1} = u_i + t_i d_i$ where $d_i$ is a directional vector with zeroes in all but one location. This method is called *dual-descent.* In both of these methods, weights are adjusted until there is no change in the relaxed solution: $S^*(u_i) = S^*(u_{i+1})$.

While better relaxed solutions will create greater reduction in the amount of subsequent refinement search, it is unclear just where the tradeoff between these two search spaces lies. Perhaps it is unnecessary to spend much time improving relaxed schedules. Thus a more radical, and extremely efficient, approach is to settle for the first relaxed solution found. We call this the *first-solution* method. A more moderate approach is to perform careful weight search at the beginning of the refinement search (where there is much to be gained by reducing the subsequent refinement search) and to perform the more restricted first-solution search when deeper in the refinement search tree. The *truncated-dual-descent* method performs dual-descent at the initial refinement search node and then uses the *first-solution* method for the rest of the refinement search.

The control grammar includes four methods for performing weight space search (Figure 3).

| | |
|---|---|
| 2a: Subgradient-optimization | 2c: Truncated-dual-descent |
| 2b: Dual-descent | 2d: First-solution |

Figure 3: Weight Search Methods

### 5.1.3 SELECT SOME CONSTRAINT

If the scheduler cannot find a relaxed solution that solves the original problem, it must break the current partial schedule into a set of refinements and explore them non-deterministically. In Adaptive LR-26, the task of creating refinements is broken into two decisions: selecting an unsatisfied constraint (Decision 3), and creating refinements that make progress towards satisfying the selected constraint (Decision 4). Lagrangian relaxation simplifies the first decision by identifying a small subset of constraints that appear problematic. However, this still leaves the problem of choosing one constraint in this subset on which to base the subsequent refinement.

The common wisdom in the search community is to choose a constraint that maximally constrains the rest of the search space, the idea being to minimize the size of the subsequent refinement search and to allow rapid pruning if the partial schedule is unsatisfiable. Therefore, our control grammar incorporates several alternative heuristic methods for locally assessing this factor. Given that the common wisdom is only a heuristic, we include a small number of methods that violate this intuition. All of these methods are functions that look at the local constraint graph topology and return a value for each constraint. Constraints can then be ranked by their value and the highest value constraint chosen. The control grammar implements both a primary and secondary sort for constraints. Constraints that have the same primary value are ordered by their secondary value.

For the sake of simplicity we only discuss measures for constraints of the form $\Sigma as \geq b$. (Analogous measures are defined for other forms.) We first define measures on time periods. Measures on constraints are functions of the measures of the time periods that participate in the constraint.

**Measures on Time Periods.** An *unforced* time period is one that is neither in or out of the schedule ($value(s)=*$). The *conflictedness* of an unforced time period $s$ (with respect to a current partial schedule) is the number of other unforced time periods that will be forced out if $s$ is forced into the schedule (because they participate in an antenna constraint with $s$). If a time period is already forced out of the current partial schedule, it does not count toward $s$'s conflictedness. Forcing a time period with high conflictedness into the schedule will result in many constraint propagations, which reduces the number of ground schedules in the refinement.

The *gain* of an unforced time period $s$ (with respect to a current partial schedule) is the number of unsatisfied project constraints that $s$ participates in. Preferring time periods with high gain will make progress towards satisfying many project constraints simultaneously.

The *loss* of an unforced time period $s$ (with respect to a current partial schedule) is a combination of gain and conflictedness. Loss is the sum of the gain of each unforced time period that will be forced out if $s$ is forced into the schedule. Time period with high loss are best avoided as they prevent progress towards satisfying many project constraints.

To illustrate these measures, consider the simplified scheduling problem in Figure 4.

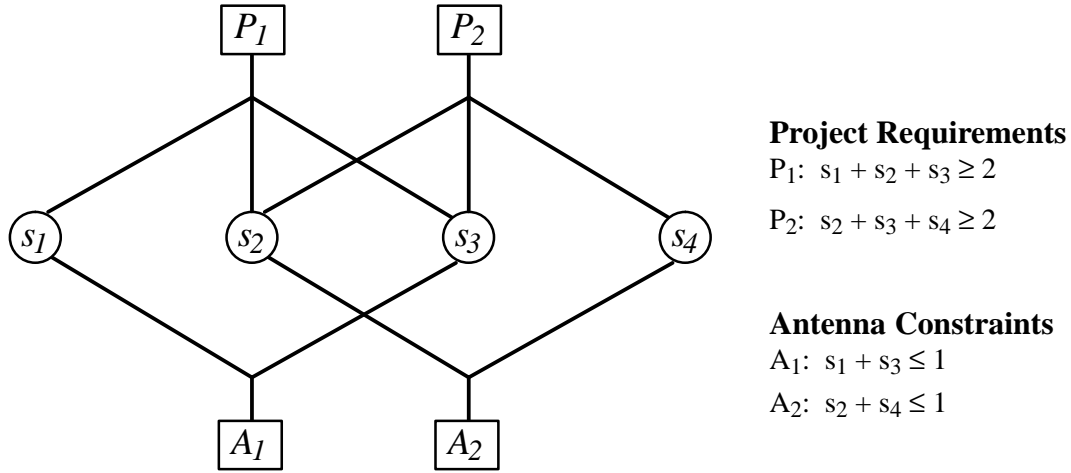

**Project Requirements**

$P_1$: $s_1 + s_2 + s_3 \geq 2$

$P_2$: $s_2 + s_3 + s_4 \geq 2$

**Antenna Constraints**

$A_1$: $s_1 + s_3 \leq 1$

$A_2$: $s_2 + s_4 \leq 1$

Figure 4: A simplified DSN scheduling problem based on four time periods. There are two project constraints, and two antenna constraints. For example, $P_1$ signifies that at least two of the first three time periods must appear in the schedule, and $A_1$ signifies that either $s_1$ or $s_3$ may appear in the schedule, but not both. In the solution, only $s_2$ and $s_3$ appear in the schedule.

With respect to the initial partial schedule (with none of the time periods forced either in or out) the conflictedness of $s_2$ is one, because it appears in just one antenna constraint ($A_2$). If subsequently, $s_4$ is forced out, then the conflictedness of $s_2$ drops to zero, as conflictedness is only computed over unforced time periods. The initial gain of $s_2$ is two, as it appears in both project constraints. Its gain drops to one if $s_3$ and $s_4$ are then forced into the schedule, as $P_2$ becomes satisfied. The initial loss of $s_2$ is the sum of the gain of all time periods conflicting with it ($s_4$). The gain of $s_4$ is one (it appears in $P_2$) so that the loss of $s_2$ is one.

**Measures on Constraints.** Constraint measures (with respect to a partial schedule) can be defined as functions of the measures of the unforced time periods that participate in a constraint. The functions *max, min,* and *total* have been defined. Thus, *total-conflictedness* is the sum of the conflictedness of all unforced time periods mentioned in a constraint, while *max-gain* is the maximum of the gains of the unforced time periods. Thus, for the constraints defined above, the initial total-conflictedness of $P_1$ is the conflictedness of $s_1$, $s_2$ and $s_3$, $1 + 1 + 1 = 3$. The initial max–gain of constraint $P_1$ is the maximum of the gains of $s_1$, $s_2$, and $s_3$ or max$\{1,2,2\} = 2$.

We also define two other constraint measures. The *unforced-periods* of a constraint (with respect to a partial schedule) is simply the number of unforced time periods that are mentioned in the constraint. Preferring a constraint with a small number of unforced time periods restricts the number of refinements that must be considered, as refinements consider combinations of time periods to force into the schedule in order to satisfy the constraint. Thus, the initial unforced-periods of $P_1$ is three ($s_1$, $s_2$, and $s_3$).

The *satisfaction-distance* of a constraint (with respect to a partial schedule) is a heuristic measure the number of time periods that must be forced in order to satisfy the constraint. The measure is heuristic because it does not account for the dependencies between time periods imposed by antenna constraints. The initial satisfaction-distance of $P_1$ is two because two time periods must be forced in before the constraint can be satisfied.

Given these constraint measures, constraints can be ordered by some measure of their worth. For example we may prefer constraints with high total conflictedness, denoted as *prefer-total-conflictedness*. Not all possible combinations seem meaningful so the control grammar for Adaptive LR-26 implements nine constraint ordering heuristics (Figure 5).

| | | |
|---|---|---|
| 3a: | Prefer-max-gain | |
| 3b: | Prefer-total-gain | |
| 3c: | Penalize-max-loss | |
| 3d: | Penalize-max-conflictedness | |
| 3e: | Prefer-total-conflictedness | |
| 3f: | Penalize-total-conflictedness |
| 3g: | Prefer-min-conflictedness |
| | 3h: Penalize-unforced-periods |
| 3i: | Penalize-satisfaction-distance |

Figure 5: Constraint Selection Methods

### 5.1.4   REFINE PARTIAL SCHEDULE

Given a selected constraint, the scheduler must create a set of refinements that make progress towards satisfying it. If the constraint is of the form $\Sigma as \geq b$ then some time periods on the left-hand-side must be forced into the schedule if the constraint is to be satisfied. Thus, refinements are constructed by identifying a set of ways to force time periods in or out of the partial schedule $s$ such that the refinements form a spanning set: $\cup\{S^i\} = S$. These refinements are then ordered and added to the agenda. Again, for simplicity we restrict discussion to constraints of form $\Sigma as \geq b$.

**The Basic Refinement Method.**   The basic method for refining a partial schedule is to take each unforced time period mentioned in the constraint and create a refinement with the time period $v_j$ forced into the schedule. Thus, for the constraints defined above, there would be three refinements to constraint P1, one with $s_1$ forced in: one with $s_2$ forced in, and one with $s_3$ forced in.

Each refinement is further refined by performing constraint propagation (arc consistency) to determine some local consequences of this new restriction. Thus, every time period that conflicts with $v_j$ is forced out of the refined partial schedule, which in turn may force other time periods to be included, and so forth. By this process, some refinements may be recognized as inconsistent (contain no ground solutions) and are pruned from the search space (for efficiency, constraint propagation is only performed when partial schedules are removed from the agenda).

Once the set of refinements has been created, they are ordered by a value ordering heuristic before being placed on the agenda. As with constraint ordering heuristics, there is a common wisdom for creating value ordering heuristics: prefer refinements that maximized the number of future options available for future assignments (Dechter & Pearl, 1987, Haralick & Elliott, 1980). The control grammar implements several heuristic methods using measures on the time periods that created the refinement. For example, one way to keep options is to prefer forcing in a time period with minimal conflictedness. As the common wisdom is only heuristic, we also incorporate a method that violates it. The control grammar includes five value ordering heuristics that are derived from the measures on time periods (Figure 6), where the last method, arbitrary, just uses the ordering of the time periods as they appear in the constraint.

| | |
|---|---|
| 1a: Prefer-gain | 1d: Prefer-conflictedness |
| 1b: Penalize-loss | 1e: Arbitrary |
| 1c: Penalize-conflictedness | |

Figure 6: Value Ordering Methods

**The Systematic Refinement Method.** The basic refinement method has one unfortunate property that may limit its effectiveness. The search resulting from this refinement method is *unsystematic* in the sense of McAllester and Rosenblitt (1991). This means that there is some redundancy in the set of refinements: $S^i \cap S^j \neq \emptyset$. Unsystematic search is inefficient in that the total size of the refinement search space will be greater than if a systematic (non-redundant) refinement method is used. This may or may not be a disadvantage in practice as scheduling complexity is driven by the size of the search space actually explored (the *effective search space*) rather than its total size. Nevertheless, there is good reason to suspect that a systematic method will lead to smaller effective search spaces.

A systematic refinement method chooses a time period that helps satisfy the selected constraint and then forms a spanning set of two refinements: one with the time period forced in and one with the time period forced out. These refinements are guaranteed to be non-overlapping. The systematic method incorporated in the control grammar uses the value ordering heuristic to choose which unforced time period to use. The two refinements are ordered based on which makes immediate progress towards satisfying the constraint (e.g., $s=1$ is first for constraints of form $\Sigma as \geq b$). The control grammar includes both the basic and systematic refinement methods (Figure 7).

| | |
|---|---|
| 4a: Basic-Refinement | 4b: Systematic-Refinement |

Figure 7: Refinement Methods

For the problem specified in Figure 4, when systematically refining constraint $P_1$, one would use the value ordering method to select among time periods $s_1$, $s_2$, and $s_3$. If $s_2$ were selected, two refinements would be proposed, one with $s_2$ forced in and one with $s_2$ forced out.

The control grammar is summarized in Figure 8. The original expert control strategy developed for LR-26 is a particular point in the control space defined by the grammar: the value ordering method is arbitrary (1e); the weight search is by dual-descent (2b); the primary constraint ordering is penalize-unforced-periods (3h); there is no secondary constraint ordering, thus this is the same as the primary ordering; and the basic refinement method is used (4a).

### 5.1.5 META-CONTROL KNOWLEDGE

The constraint grammar defines a space of close to three thousand possible control strategies. The quality of a strategy must be assessed with respect to a distribution of problems, therefore it is prohibitively expensive to exhaustively explore the control space: taking a significant number of examples (say fifty) on each of the strategies at a cost of 5 CPU minutes per problem would require approximately 450 CPU days of effort.

CONTROL STRATEGY :=        VALUE ORDERING ∧
                           WEIGHT SEARCH METHOD ∧
                           PRIMARY CONSTRAINT ORDERING  ∧
                           SECONDARY CONSTRAINT ORDERING ∧
                           REFINEMENT METHOD

| | |
|---|---|
| VALUE ORDERING | := {1a, 1b, 1c, 1d,1e} |
| WEIGHT SEARCH METHOD | := {2a, 2b, 2c, 2d} |
| PRIMARY CONSTRAINT ORDERING | := {3a, 3b, 3c, 3d, 3e, 3f, 3g, 3h, 3i} |
| SECONDARY CONSTRAINT ORDERING | := {3a, 3b, 3c, 3d, 3e, 3f, 3g, 3h, 3i} |
| REFINEMENT METHOD | := {4a, 4b} |

Figure 8: Control grammar for Adaptive LR-26

COMPOSER requires a transformation generator to specify alternative strategies, which are explored via hillclimbing search. In this case, the obvious way to proceed is to consider all single method changes to a given control strategy. However the cost of searching the strategy space and quality of the final solution depend to a large extent on how hillclimbing proceeds, and the obvious way need not be the best. In Adaptive LR-26, we augment the control grammar with some domain-specific knowledge to help organize the search. Such knowledge includes, for example, our prior expectation that certain control decisions would interact, and the likely importance of the different control decisions. The intent of this "meta-control knowledge" is to reduce the branching factor in the control strategy search and improve the expected utility of the locally optimal solution found. This approach led to a layered search through the strategy space. Each control decision is assigned to a level. The control grammar is search by evaluating all combinations of methods at a single level, adopting the best combinations, and then moving onto the next level. The organization is shown below:

Level 0:     {Weight search method}
Level 1:     {Refinement method}
Level 2:     {Secondary constraint ordering, Value ordering}
Level 3:     {Primary constraint ordering}

The weight search and refinement control points are separate, as they seem relatively independent from the other control points, in terms of their effect on the overall strategy. While there is clearly some interaction between weight search, refinement construction, and the other control points, a good selection of methods for pricing and alternative construction should perform well across all ordering heuristics. The primary constraint ordering method is relegated to the last level because some effort was made in optimizing this decision in the expert strategy for LR-26, and we believed that it was unlikely the default strategy could be improved.

Given this transformation generator, Adaptive LR-26 performs hillclimbing across these levels. It first entertains weight adjustment methods, then alternative construction methods, then combinations of secondary constraint sort and child sort methods, and finally primary constraint sort methods. Each choice is made given the previously adopted methods.

This layered search can be viewed as the consequence of asserting certain types of relations between control points. *Independence relations* indicate cases in which the utility of methods for one control point is roughly independent of the methods used at other control points. *Dominance rela-* *tions* indicate that the changes in utility from changing methods for one control point are much larger than the changes in utility for another control point. Finally, inconsistency relations indicate when a method $M_1$ for control point $X$ is inconsistent with method $M_2$ for control point $Y$. This means that any strategy using these methods for these control points need not be considered.

## 5.2 EXPECTED UTILITY

As previously mentioned, a chief design requirement for LR-26 is that the scheduler produce solutions (or prove that none exist) efficiently. This behavioral preference can be expressed by a utility function related to the computational effort required to solve a problem. As the effort to produce a schedule increases, the utility of the scheduler on that problem should decrease. In this paper, we characterize this preference by defining utility as the negative of the CPU time required by the scheduler on a problem. Thus, Adaptive LR-26 tunes itself to strategies that minimize the average time to generate a schedule (or prove that one does not exist). Other utility functions could be entertained. In fact, more recent research has focused on measures of schedule quality (Chien & Gratch, 1994).

## 5.3 Problem Distribution

Adaptive LR-26 needs a representative sample of training examples for its adaptation phase. Unfortunately, DSN Operations has only recently begun to maintain a database of scheduling problems in a machine readable format. While this will ultimately allow the scheduler to tune itself to the actual problem distribution, only a small body of actual problems was available at the time of this evaluation. Therefore, we resorted to other means to create a reasonable problem distribution.

We constructed an augmented set of training problems by syntactic manipulation of the set of real problems. Recall that each scheduling problem is composed of two components: a set of project requirements, and a set of time periods. Only the time periods change across scheduling problems, so we can organize the real problems into a set of tuples, one for each project, containing the weekly blocks of time periods associated with it (one entry for each week the project is scheduled). The set of augmented scheduling problems is constructed by taking the cross product of these tuples. Thus, a weekly scheduling problem is defined by combining one weeks worth of time periods from each project (time periods for different projects may be drawn from different weeks), as well as the project requirements for each. This simple procedure defines set of 6600 potential scheduling problems.

Two concerns led us to use only a subset of these augmented problems. First, a significant percentage of augmented problems appeared much harder to solve (or prove unsatisfiable) than any of the real problems (on almost half of the constructed problems the scheduler did not terminate, even with large resource bounds). That such "hard" problems exist is not unexpected as scheduling is NP-hard, however, their frequency in the augmented sample seems disproportionately high. Second, the existence of these hard problems raises a secondary issue of how best to terminate search. The standard approach is to impose some arbitrary resource bound and to declare a problem unsatisfiable if no solution is found within this bound. Unfortunately this raises the issue of what sized bound is most reasonable. We could have resolved this by adding the resource bound to the control grammar, however, at this point in the project we settled for a simpler approach. We address this and the previous concern by excluding from the augmented problem distribution those problems that seem "fundamentally intractable." What this means in practice is that we exclude problems that could not be solved by any of a large set of heuristic methods within a five minute resource bound, the determina-tion of which is discussed in Appendix A. This results in a reduced set of about three thousand scheduling problems.

The use of a resource bound can be problematic for evaluating the power of a learning technique. As noted by Segre, Elkan, and Russell (1991), a learning system that greatly improves problem solving performance under a given resource bound may perform quite differently under a different resource bound. Some researchers suggest statistical analysis methods for assessing the significance of this factor (e.g., see Etzioni and Etzioni, 1994). In this study, however, we do not address the issue of how results might change given different resource bounds. We note that COMPOSER's statistical properties suggest that problem solving performance should be no worse after learning, whatever the resource bound, but the performance improvement many vary considerably. To give at least some insight into the generality of adaptive problem solving, we include a secondary set of evaluations based on all 6600 augmented problems (including fundamentally "intractable" ones).

## 6. Empirical Evaluation

We conjecture that Adaptive LR–26 will improve the performance of the basic scheduler. This can be broken down into two separate claims. First, we claim that the modifications suggested above contain useful transformations (it is possible to improve the scheduler). Second, we claim that Adaptive LR–26 should identify these transformations (and avoid harmful ones) with the requested level of probability. The first claim is solely based on our intuitions; the second supported by the statistical theory that underlies the COMPOSER approach. The usefulness of COMPOSER depends on its ability to COMPOSER can go beyond simply improving performance and identifying strategies that rank highly when judged with respect to the whole space of possible strategies. A third claim, therefore, is that Adaptive LR-26 will find better strategies than if we simply picked the best of a large number of randomly selected strategies. Besides testing these three claims, we are also interested in three secondary questions: how quickly does the technique improve expected utility (e.g., how many examples are required to make statistical inferences?); can Adaptive LR-26 improve the number problems solved (or proved unsatisfiable) within the resource bound; and how sensitive is the effectiveness of adaptive problem solving to changes in the distribution of problems.

### 6.1 Methodology

Our evaluation is influenced by the stochastic nature of adaptive problem solving. During adaptation, Adaptive LR-26 is guided by a random selection of training examples according to the problem distribution. As a result of this random factor, the system will exhibit different behavior on different runs of the system. On some runs the system may learn high utility strategies; on other runs the random examples may poorly represent the distribution and the system may adopt transformations with negative utility. Thus, our evaluation is directed at assessing the *expected* performance of the adaptive scheduler by averaging results over multiple experimental trials.

For these experiments, the scheduler is allowed to adapt to 300 scheduling problems drawn randomly from the problem distribution described above. The expected utility of all learned strategies is assessed on an independent test set of 1000 test examples drawn randomly from the complete set of three thousand. The adaptation rate is assessed by recording the strategy learned by Adaptive LR-26 after every 20 examples. Thus we can see the result of learning with only twenty examples, only forty examples, etc. We measure the statistical error of the technique (the probability of adopting a trans-

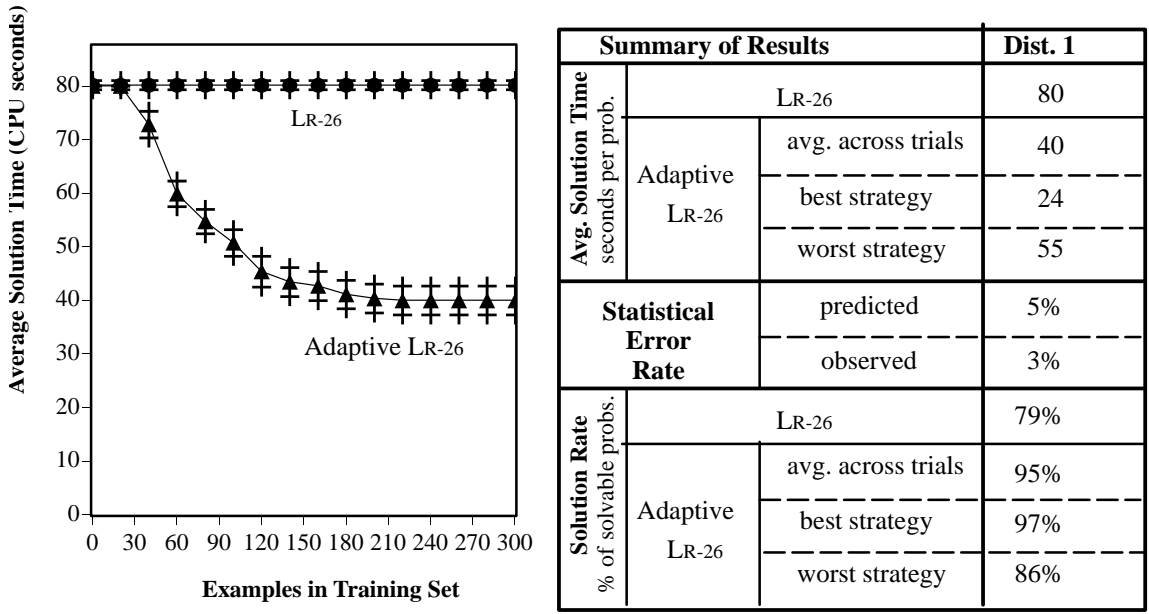

Figure 9. Learning curve showing performance as a function of the number of training examples and table of experimental results.

formation with negative incremental utility) by performing eighty runs of the system on eighty distinct training sets drawn randomly from the problem distribution. We measure the distributional sensitivity of the technique by evaluating the adaptive scheduler on a second distribution of problems. Recall that we purposely excluded inherently difficult scheduling problems from the augmented set of problems. If added, these excluded problems should make adaptation more difficult as no strategy is likely to provide a noticeable improvement within the five minute resource bound. The second evaluation includes these difficult problems

A third evaluation assesses the relative quality of the strategies identified by Adaptive LR-26 when compared with other strategies in the strategy space. This is inferred by comparing the expected utility of the learned strategies with several strategies drawn randomly from the space. This also provides an opportunity to assess the quality of the expert strategy, and thus give a sense of how challenging it is to improve it.

COMPOSER, the statistical component of the adaptive scheduler, has two parameters that govern its behavior. The parameter $\delta$ specifies the acceptable level of statistical error (this is the chance that the technique will adopt a bad transformation or reject a good one). In Adaptive LR-26, this is set to a standard value of 5%. COMPOSER bases each statistical inferences on a minimum of $n_0$ examples. In Adaptive LR-26, $n_0$ is set to the empirically determined value of fifteen.

## 6.2 Overall Results — DSN DISTRIBUTION

Figure 9 summarizes the results of adaptive problem solving over the constructed DSN problem distribution. The results support the two primary claims. First, the system learned search control strategies that yielded a significant improvement in performance. Adaptive problem solving reduced the average time to solve a problem (or prove it unsatisfiable) from 80 to 40 seconds (a 50% improvement). Second, the observed statistical error fell well within the predicted bound. Of the 370 transformations adopted across the eighty trials, only 3% decreased expected utility.

Due to the stochastic nature of the adaptive scheduler, different strategies were learned on different trials. All learned strategies produced at least some improvement in performance. The best of these strategies required only 24 seconds on average to solve a problem (an improvement of 70%). The fastest adaptations occurred early in the adaptation phase and performance improvements decreased steadily throughout. It took an average of 62 examples to adopt each transformation. Adaptive LR-26 showed some improvement over the non-adaptive scheduler in terms of the number of problems that could be solved (or proven unsatisfiable) within the resource bound. LR-26 was unable to solve 21% of the scheduling problems within the resource bound. One adaptive strategy substantially reduced this number to 3%.

An analysis of the learned strategies is revealing. Most of the performance improvement (about one half) can be traced to modifications in LR-26's weight search method. The rest of the improvements are divided equally among changes to the heuristics for value ordering, constraint selection, and refinement. As expected, changes to the primary constraint ordering only degraded performance. The top three strategies are illustrated in Figure 10.

| | | |
|---|---|---|
| 1) | Value ordering: | penalize-conflictedness (1c) |
| | Weight search: | first-solution (2d) |
| | Primary constraint ordering: | penalize-unforced-periods (3h) |
| | Secondary constraint ordering: | prefer-total-conflictedness (3e) |
| | Refinement method: | systematic-refinement (4b) |
| | | |
| 2) | Value ordering: | prefer-gain (1a) |
| | Weight search: | first-solution (2d) |
| | Primary constraint ordering: | penalize-unforced-periods (3h) |
| | Secondary constraint ordering: | prefer-total-conflictedness (3e) |
| | Refinement method: | systematic-refinement (4b) |
| | | |
| 3) | Value ordering: | penalize-conflictedness (1c) |
| | Weight search: | first-solution (2d) |
| | Primary constraint ordering: | penalize-unforced-periods (3h) |
| | Secondary constraint ordering: | penalize-satisfaction-distance (3i) |
| | Refinement method: | systematic-refinement (4b) |

Figure 10: The three highest utility strategies learned by Adaptive LR-26.

For the weight search, all of the learned strategies used the first-solution method (2d). It seems that, at least in this domain and problem distribution, the reduction in refinement search space that results from better relaxed solutions is more than offset by the additional cost of the weight search. The scheduler did, however, benefit from the reduction in size that results from a systematic refinement method.

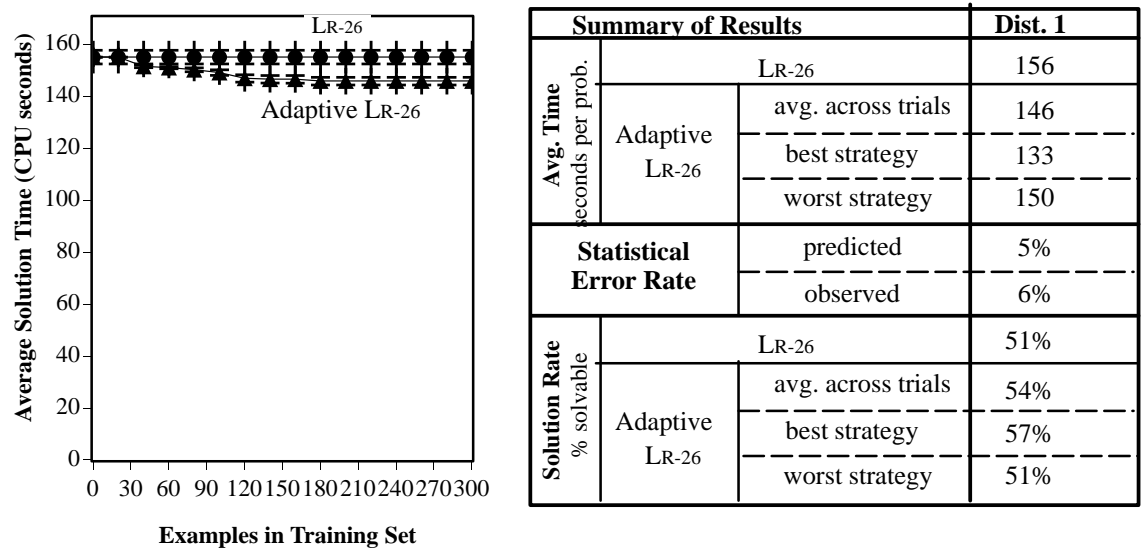

Figure 11. Learning curves and table of experimental results showing performance over the augmented distribution (including "intractable" problems).

More interestingly, Adaptive L$_R$-26 seems to have "rediscovered" the common wisdom in heuristic constraint-satisfaction search. When exploring new refinements, it is often suggested to chose the least constrained value of the most constrained constraint. The best learned strategies follow this advice while the worst strategies violate it. In the best strategy, the time period with lowest conflictedness is least constraining (in the sense that it will tend to produce the least constraint propagations) and thus produces the least commitments on the resulting partial schedule. By this same argument, the constraint with the highest total conflicted will tend to be the hardest to satisfy.

## 6.3 Overall Results — FULL AUGMENTED DISTRIBUTION

Figure 11 summarizes the results for the augmented distribution. As expected, this distribution proved more challenging for adaptive problem solving. Nevertheless, modest performance improvements were still possible, lending support to our claimed generality of the adaptive problem solving approach. Learned strategies reduced the average solution time from 156 to 146 seconds (an 6% improvement). The best learned strategies required 133 seconds on average to solve a problem (an improvement of 15%). The observed statistical accuracy did not significantly differ from the theoretically predicted bound, although it was slightly higher than expected: of 397 transformations were adopted across the trials, 6% produced a decrease in expected utility. The introduction of the difficult problems resulted in higher variance in the distribution of incremental utility values and this is reflected in a higher sample complexity: an average of 118 examples to adopt each transformation. Some improvement was noted on the supposedly intractable problems. One strategy learned by Adaptive L$_R$-26 increased the number of problems that could be processed within the resource bound from 51% to 57%.

One interesting result of this evaluation is that, unlike the previous evaluation, the best learned strategies use truncated-dual-descent as their weight search method (the strategies were similar along other control dimensions). This illustrates how even modest changes to the distribution of problems can influence the design tradeoffs associated with a problem solver: in this case, changing the tradeoff between weight and refinement search.

## 6.4 Quality of Learned strategies

The third claim is that, in practice, COMPOSER can identify strategies that rank highly when judged with respect to the whole strategy space. A secondary question is how well does the expert strategy perform. The improvements of Adaptive LR-26 are of little significance if the expert strategy performs worse than most strategies in the space. Alternatively, if the expert strategy is extremely good, its improvement is compelling.

As a way of assessing these claims we estimate the probability of selecting a high utility strategy given that we choose it randomly from one of three strategy spaces: the space of all possible strategies (expressible in the transformation grammar), the space of strategies produced by Adaptive LR-26, and the trivial space containing only the expert strategy. This corresponds to the problem of estimating a *probability density function* (p.d.f.) for each space: a p.d.f., $f(x)$, associated with a random variable gives the probability that an instance of the variable has value $x$. More specifically we want to estimate the density functions, $f_s(u)$, which is the probability of randomly selecting a strategy from space $s$ that has expected utility $u$.

We use a non-parametric density estimation technique called the kernel method to estimate $f_s(u)$ (as in Smyth, 1993). To estimate the density function of the whole space, we randomly selected and tested thirty strategies. All of the learned strategies are used to estimate the density of the learned space. (In both cases, five percent of the data was withheld to estimate the bandwidth parameter used by the kernel method.) The p.d.f. associated with the single expert strategy is estimated using a normal model fit to the 1000 test examples from the previous evaluation.

### 6.4.1 DSN DISTRIBUTION

Figure 12 illustrates the results for the DSN distribution. In this evaluation the learned strategies significantly outperformed the randomly selected strategies. Thus, one would have to select and test many strategies at random before finding one of comparable expected utility to one found by Adaptive LR-26. The results also indicate that the expert strategy is already a good strategy (as indicated by the relative positions of the peaks for the expert and random strategy distributions), indicating that the improvement due to Adaptive LR-26 is significant and non-trivial.

The results provide additional insight into Adaptive LR-26's learning behavior. That the p.d.f for the learned strategies contains several peaks, graphically illustrates that different local maxima exist for this problem. Thus, there may be benefit in running the system multiple times and choosing the best strategy. It also suggests that techniques designed to avoid local maxima would be beneficial.

### 6.4.2 FULL AUGMENTED DISTRIBUTION

Figure 13 illustrates the results for the full augmented distribution. The results are similar to the DSN distribution: the learned strategies again outperformed the expert strategy which in turn again outperformed the randomly selected strategies. The data shows that the expert strategy is significantly better than randomly selected strategies. Together, these two evaluations support the claim that Adaptive LR-26 is selecting high performance strategies. Even though the expert strategy is quite good when compared with the complete strategy space, the adaptive algorithm is able to improve the expected problem solving performance.

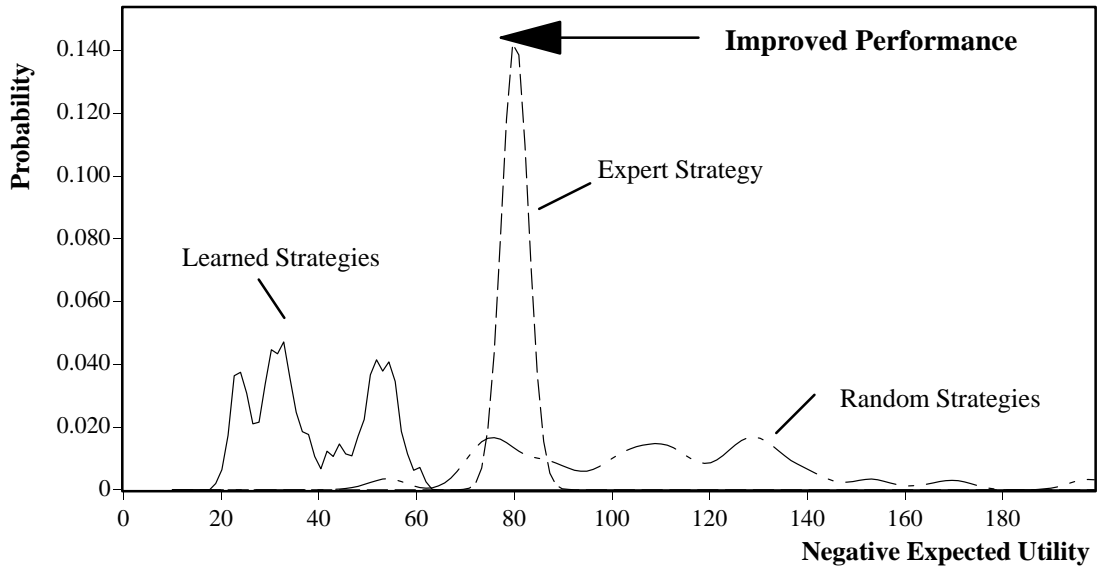

Figure 12: The DSN Distribution. The graph shows the probability of obtaining a strategy of a particular utility, given that it is chosen from (1) the set of all strategies, (2) the set of learned strategies, or (3) the expert strategy.

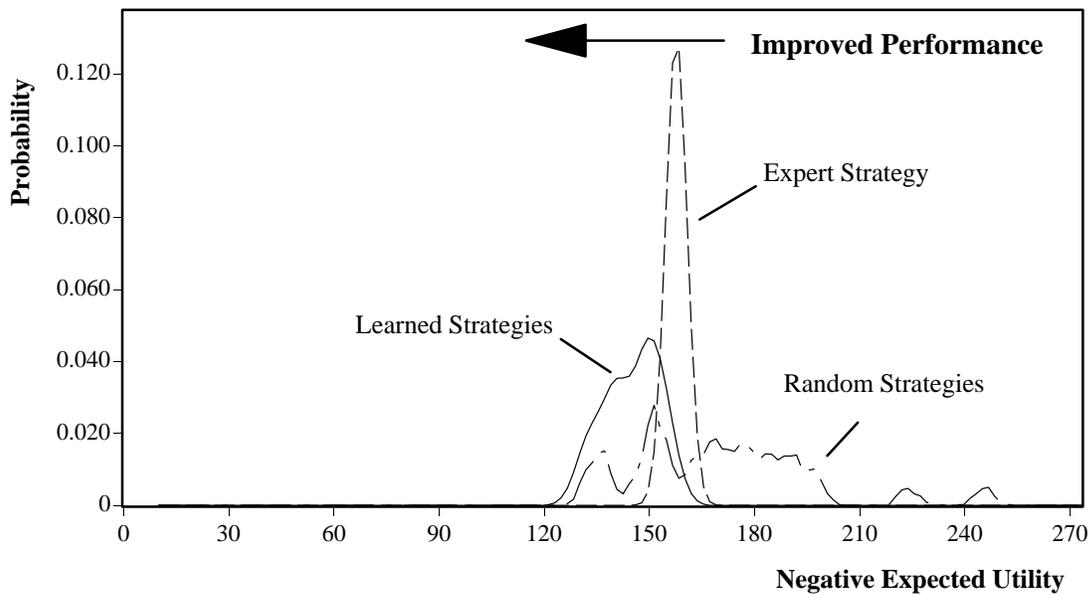

Figure 13: The Full augmented distribution. The graph shows the probability of obtaining a strategy of a particular utility, given that it is chosen from (1) the set of all strategies, (2) the set of learned strategies, or (3) the expert strategy.

## 7.    Future Work

The results of applying an adaptive approach to deep space network scheduling are very promising. We hope to build on this success in a number of ways. We discuss these directions as they relate to the three basic approaches to adaptive problem solving: syntactic, generative, and statistical.

### 7.1  Syntactic Approaches

Syntactic approaches attempt to identify control strategies by analyzing the structure of the domain and problem solver. In LR-26, our use of meta-control knowledge can be seen as a syntactic approach; although unlike most syntactic approaches that attempt to identify a specific combination of heuristic methods, the meta-knowledge (dominance and indifference relations) acts as constraints that only partially determine a strategy. An advantage of this weakening of the syntactic approach is that it lends itself to a natural and complementary interaction with statistical approaches: structural information restricts the space of reasonable strategies, which is then explored by statistical techniques. An important question concerning such knowledge is to what extent does it contribute to the success of our evaluations, and, more interestingly, how could such information be derived automatically from a structural analysis of the domain and problem solver. We are currently performing a series of experiments to address the former question. A step towards the resolving the second question would be to evaluate in the context of LR-26 some of the structural relationships suggested by recent work in this area (Frost & Dechter, 1994, Stone, Veloso & Blythe, 1994).

### 7.2  Generative Approaches

Adaptive LR-26 uses a non-generative approach to conjecturing heuristics. Our experience in the scheduling domain indicates that the performance of adaptive problem solving is inextricably tied to the transformations it is given and the expense of processing examples. Just as an inductive learning technique relies on good attributes, if COMPOSER is to be effective, there must exist good methods for the control points that make up a strategy. Generative approaches could improve the effectiveness of Adaptive LR-26. Generative approaches dynamically construct heuristic methods in response to observed problem-solving inefficiencies. The advantage of waiting until inefficiencies are observed is twofold. First, the exploration of the strategy space can be much more focused by only conjecturing heuristics relevant to the observed complications. Second, the conjectured heuristics can be tailored much more specifically to the characteristics of these observed complications.

Our previous application of COMPOSER achieved greater performance improvements than Adaptive LR-26, in part because it exploited a generative technique to construct heuristics (Gratch & DeJong, 1992). Ongoing research is directed towards incorporating generative methods into Adaptive LR-26. Some preliminary work analyzes problem-solving traces to induce good heuristic methods. The constraint and value ordering metrics discussed in Section 5.1.3 are used to characterize each search node. This information is then fed to a decision-tree algorithm, which tries to induce effective heuristic methods. These generated methods can then be evaluated statistically.

### 7.3  Statistical Approaches

Finally there are directions of future work devoted towards enhancing the power of the basic statistical approach, both for Adaptive LR-26 in particular, and for statistical approaches in general.

For the scheduler, there are two important considerations: enhancing the control grammar and exploring a wider class of utility functions. Several methods could be added to the control grammar. For example, an informal analysis of the empirical evaluations suggests that the scheduler could benefit from a look-back scheme such as backjumping (Gaschnig, 1979) or backmarking (Haralick & Elliott, 1980). We would also like to investigate the adaptive problem solving methodology on a richer variety of scheduling approaches, besides integer programming. Among these would be more powerful bottleneck centered techniques (Biefeld & Cooper, 1991), constraint-based techniques (Smith & Cheng, 1993), opportunistic techniques (Sadeh, 1994), reactive techniques (Smith, 1994) and more powerful backtracking techniques (Xiong, Sadeh & Sycara, 1992).

The current evaluation of the scheduler focused on problem solving time as a utility metric, but future work will consider how to improve other aspects of the schedulers capabilities. For example, by choosing another utility function we could guide Adaptive LR-26 towards influencing other aspects of LR-26's behavior such as: increasing the amount of flexibility in the generated schedules, increasing the robustness of generated schedules, maximizing the number of satisfied project constraints, or reducing the implementation cost of generated schedules. These alternative utility functions are of great significance in that they provide much greater leverage in impacting actual operations. For example, finding heuristics which will reduce DSN schedule implementation costs by 3% would have a much greater impact than reducing the automated scheduler response time by 3%. Some preliminary work has focused on improving schedule quality (Chien & Gratch, 1994).

More generally, there are several ways to improve the statistical approach embodied by COMPOS-ER. Statistical approaches involve two processes, estimating the utility of transformations and exploring the space of strategies. The process of estimating expected utilities can be enhanced by more efficient statistical methods (Chien, Gratch & Burl, 1995, Moore & Lee, 1994, Nelson & Matejcik, 1995), alternative statistical decision requirements (Chien, Gratch & Burl, 1995) and more complex statistical models that weaken the assumption of normality (Smyth & Mellstrom, 1992). The process of exploring the strategy space can be improved both in terms of its efficiency and susceptibility to local maxima. Moore and Lee propose a method called *schemata search* to help reduce the combinatorics of the search. Problems with local maxima can be mitigated, albeit expensively, by considering all *k*-wise combinations of heuristics (as in MULTI-TAC or level 2 of Adaptive LR-26's search), or by standard numerical optimization approaches such as repeating the hillclimbing search several times from different start points.

One final issue is the expense in processing training examples. In the LR-26 domain this cost grows linearly with the number of candidates at each hillclimbing step. While this is not bad from a complexity standpoint, it is a pragmatic concern. There have been a few proposals to reduce the expense in gathering statistics. In previous work (Gratch & DeJong, 1992) we exploited properties of the transformations to gather statistics from a single solution attempt. That system required that the heuristic methods only act by pruning refinements that are guaranteed unsatisfiable. Greiner and Jurisica (1992) discuss a similar technique that eliminates this restriction by providing upper and lower bounds on the incremental utility of transformations. Unfortunately, neither of these approaches could be applied to LR-26 so devising methods to reduce the processing cost is an important direction for future work.

## 8. Conclusions

Although many scheduling problems are intractable, for actual sets of constraints and problem distributions, heuristic solutions can provide acceptable performance. A frequent difficulty is that determining appropriate heuristic methods for a given problem class and distribution is a challenging process that draws upon deep knowledge of the domain and the problem solver used. Furthermore, if the problem distribution changes some time in the future, one must manually re-evaluate the effectiveness of the heuristics.

Adaptive problem solving is a general approach for reducing this developmental burden. This paper has described the application of adaptive problem solving, using the LR–26 scheduling system and the COMPOSER machine learning system, to automatically learn effective scheduling heuristics for Deep Space Network communications scheduling. By demonstrating the application of these techniques to a real-world application problem, this paper has makes several contributions. First, it provides an example of how a wide range of heuristics can be integrated into a flexible problem-solving architecture — providing an adaptive problem-solving system with a rich control space to search. Second, it demonstrates that the difficulties of local maxima and large search spaces entailed by the rich control space can be tractably explored. Third, the successful application of the COMPOSER statistical techniques demonstrates the real-world applicability of the statistical assumptions underlying the COMPOSER approach. Fourth, and most significantly, this paper demonstrates the viability of adaptive problem solving. The strategies learned by the adaptive problem solving significantly outperformed the best human expert derived solution.

## Appendix A. Determination of the Resource bound

A good CPU bound to characterize "intractable" problems should have the characteristic that increasing the bound should have little effect on the proportion of problems solvable. In order to determine the resource bound to define "intractable" DSN scheduling problems we empirically evaluated how likely LR–26 was to be able to solve a problem with various resource bounds. Informally, we experimented to find a bound of 5 CPU minutes. We then formally verified this bound by taking those problems not solvable within the resource bound of 5 CPU minutes, allowing LR–26 an additional CPU hour to attempt to solve the problem, and observing how this affected solution rate. As expected, even allocating significant more CPU time, LR–26 was not able to solve many more problems. Figure 14 below shows the cumulative percentage of problems solved; from those not solvable within the 5 minute CPU bound. This curve shows that even with another CPU hour (per problem!), only about 12% of the problems became solvable. This graph also shows the 95% confidence intervals for this cumulative curve. In light of these results, the fact that one learned strategy was able to increase by 18% the percentage of problems solvable within the resource bound is even more impressive. In effect, learning this strategy has a greater impact than allocating another CPU hour per problem.

## Acknowledgements

Portions of this work were performed by the Jet Propulsion Laboratory, California Institute of Technology, under contract with the National Aeronautics and Space Administration and portions at the Beckman Institute, University of Illinois under National Science Foundation Grant NSF–IRI–92–09394.

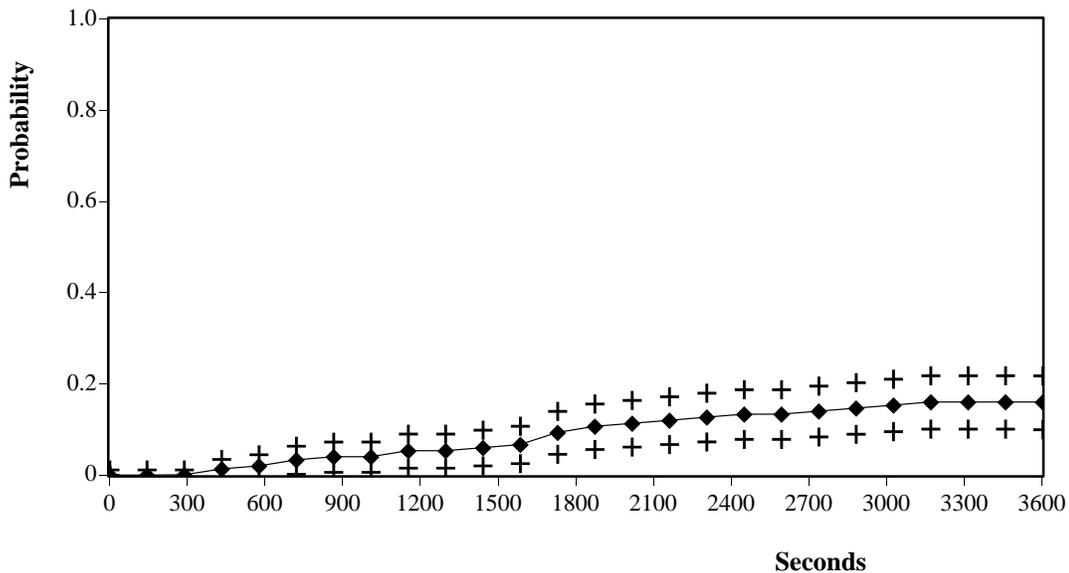

Figure 14: Given that a problem cannot be solved in five minutes, show the probability that it can be solved in up to an hour more time (with 95% confidence intervals).